%% file: paper_arxiv.tex
\documentclass[10pt,twocolumn,letterpaper]{article}

\usepackage{wacv}                            % FINAL mode: no [review] -> no anonymisation, no line numbers

\usepackage{amsmath,amssymb}
\usepackage{booktabs}
\usepackage{multirow}
\usepackage{tikz}
\usetikzlibrary{arrows.meta,positioning,shapes.geometric}
\usepackage{graphicx}
\usepackage{xspace}

\definecolor{wacvblue}{rgb}{0.21,0.49,0.74}
\usepackage[pagebackref,breaklinks,colorlinks,allcolors=wacvblue]{hyperref}

\def\wacvPaperID{}
\def\confName{}
\def\confYear{}

\def\method{EquiGrad-CAM}
\def\methodpcf{EquiGrad-CAM+PCF}

\title{Signal or Noise? Auditing Rotation-Induced Saliency Drift\\in Medical and Aerial Imaging}

\author{Khawaja Murad ul Hassan\\
Independent Researcher\\
Islamabad, Pakistan\\
{\tt\small khawajamurad@outlook.com}
\and
Mehran Ebrahimi\\
Ontario Tech University\\
Oshawa, Ontario, Canada\\
{\tt\small mehran.ebrahimi@ontariotechu.ca}
}

\begin{document}
\maketitle

\input{sec/0_abstract}
\input{sec/1_intro}
\input{sec/2_related}
\input{sec/4_signalnoise}
\input{sec/3_method}
\input{sec/6_application}
\input{sec/5_imagenet}
\input{sec/7_conclusion}

\section*{Acknowledgements}
This research was enabled in part by support provided by the Digital Research Alliance of
Canada (\url{https://alliancecan.ca}). Computations were performed on the Narval cluster.

{
    \small
    \bibliographystyle{ieeenat_fullname}
    \bibliography{refs}
}

\end{document}

%% file: sec/0_abstract.tex
\begin{abstract}
Post-hoc saliency maps such as Grad-CAM are increasingly used to \emph{audit} why a
deployed vision model made a decision, yet the heatmap drifts when the input is
rotated, even when the prediction is unchanged. In domains with no canonical
orientation, such as histopathology and aerial imagery, this drift directly undermines
using saliency as evidence. We take the drift itself as a question: when an explanation
changes under rotation, is that faithful \emph{signal} (the network genuinely attending
elsewhere) or \emph{operator noise} introduced by the CAM operator? We answer it by opening the operator
and measuring equivariance at each of its stages, rather than inferring it from the
network's output. The instability is not where one would guess: the channel weights are the
\emph{most} rotation-stable stage, and on ResNet-50 exactly stable, because a GAP+linear
head makes the class gradient field spatially constant. What moves is the spatial
activation tensor, and the classifier's own pooling discards that movement. A causal test
confirms the consequence: occluding the pixels whose saliency drifts costs the model
\emph{less} than occluding random pixels, at either orientation. The drift is therefore
carried by degrees of freedom the classifier throws away, which is what makes removing it
faithful rather than destructive. \method{} is a training-free
wrapper that takes $T$ rotated views, inverse-rotates each view's saliency back to a
common canonical frame, and averages, making a pretrained Grad-CAM (CNN or ViT)
approximately rotationally equivariant. On the full ImageNet-1K validation set it raises
equivariance over single-view Grad-CAM by $+36.0\%$ (ResNet-50), $+87.5\%$ (VGG-16), and
$+247\%$ (ViT-B/16); a scale-matched ablation isolates canonical-frame \emph{alignment
before averaging} (not where the averaging happens) as the driver, with unaligned
averaging falling to or below single-view Grad-CAM on the CNN backbones. Rotation augmentation during training
only partially closes the gap and costs clean accuracy, whereas our wrapper needs no
retraining. We demonstrate it as a saliency-auditing tool on two rotation-natural
domains, histopathology (PatchCamelyon) and aerial scene classification (RESISC45),
where it produces rotation-consistent explanations training-free, and on zero-shot CLIP,
where it lifts equivariance $+145\%$ while \emph{improving} both faithfulness measures. Its by-product,
PEUM (a per-image explanation-uncertainty score), needs no computation beyond the views the
wrapper has already taken and turns into a review-budget triage rule: at
a $10\%$ inspection budget it surfaces $4.4\times$ (aerial) and $2.8\times$
(histopathology) more rotation-unstable explanations than random inspection, giving a human
auditor a concrete place to look. Code is available at
\url{https://github.com/Khawaja-Murad/EquiGrad-CAM}.
\end{abstract}

%% file: sec/1_intro.tex
\section{Introduction}
\label{sec:intro}

Post-hoc saliency maps are increasingly used to \emph{audit} deployed vision models, to
check \emph{why} a network made a decision before that decision is trusted in a
safety- or compliance-critical setting~\cite{gradcam,euaiact,cose}. Grad-CAM~\cite{gradcam}
is the default tool: one forward--backward pass yields a heatmap over the input. Yet
Grad-CAM and its variants share a limitation that is fatal for auditing: \textbf{the
heatmap is conditioned on a single geometric configuration of the input}. Rotate the image
by $30^\circ$ and the map changes qualitatively, often while the prediction does not.

This is not a corner case. In many high-value domains the input has \emph{no canonical
orientation} at all: a pathologist rotates a slide freely, and an aerial tile has no
intrinsic ``up.'' There, two rotations of one image are equally valid views of the same
evidence, so an explanation that changes between them cannot be used as evidence: the
auditor has no principled way to decide which map to believe~\cite{cose}.

\noindent\textbf{A question the drift itself poses.} A natural objection precedes any fix:
perhaps the map \emph{should} change, because a rotated (partly out-of-distribution) input
makes the network genuinely compute something different, and forcing consistency would hide
a faithful change. We take this objection as our central question rather than an
assumption. When an explanation drifts under rotation, is that faithful \emph{signal} (the
network attending elsewhere) or \emph{operator noise} introduced by the CAM operator? We
answer it empirically (\S\ref{sec:signalnoise}): under rotation the network's internal
representation is nearly stable (feature/logit cosine $0.90/0.85$), the drift persists on
the $74\%$ of rotations that leave the prediction unchanged, and it is uncorrelated with
confidence change, so the drift is \emph{mostly operator noise}, and removing it recovers
the explanation the model's own stable state already implies.

\noindent\textbf{Removing the noise, training-free.} Averaging over rotated views is the
natural remedy, and is standard for \emph{predictions}~\cite{resnet}. But the obvious
instantiation (averaging the finished heatmaps, as multi-view saliency methods such as
Augmented Score-CAM~\cite{augscam} do) is not enough, and on fragile backbones it
\emph{degrades} equivariance below the single-view baseline. The missing ingredient is
\textbf{alignment}: each view's saliency must be inverse-rotated back to a common canonical
frame \emph{before} averaging, so that consistent evidence reinforces and operator noise
cancels. \method{} does exactly this: $T$ rotated views, each inverse-aligned, then
averaged, as a training-free wrapper around any pretrained Grad-CAM, CNN or ViT. A
scale-matched ablation (\S\ref{sec:experiments}) shows that this canonical-frame alignment,
\emph{not} where the averaging happens (feature space vs.\ aligned output space, which are
near-identical at full scale), is what drives the gain.

\noindent\textbf{Auditing rotation-natural domains.} We frame the wrapper as a
saliency-auditing tool and demonstrate it where it matters most: two rotation-natural
domains, histopathology (PatchCamelyon) and aerial scene classification (RESISC45)
(\S\ref{sec:application}). On both, single-view Grad-CAM drifts sharply under rotation, our
wrapper produces rotation-consistent explanations without retraining, and its by-product
PEUM (a per-image explanation-uncertainty score) ranks explanations by how reproducible
they are, telling an auditor which maps not to trust unexamined.

\noindent\textbf{Contributions.}
\begin{enumerate}\itemsep2pt
\item \textbf{A question, answered by decomposing the operator.} We measure equivariance at
every stage of Grad-CAM rather than inferring it from the network's output, and find the
drift is carried by degrees of freedom the classifier discards: the channel weights
$\alpha$ are the \emph{most} rotation-stable stage, on ResNet-50 exactly so because a
GAP+linear head makes the gradient field spatially constant, while the spatial
activations move, and the head pools that movement away
(\S\ref{sec:signalnoise}). A causal test closes the argument: occluding the drifting
pixels costs the model less than occluding random ones at either orientation. This
justifies removing the drift and addresses the main objection to consistency-based
explanations.
\item \textbf{\method{}, a training-free equivariant wrapper.} Canonical-frame-aligned
multi-view aggregation improves equivariance over Grad-CAM by $+36.0\%$ (ResNet-50),
$+87.5\%$ (VGG-16) and $+247\%$ (ViT-B/16) at full ImageNet-1K scale. A scale-matched
ablation isolates \emph{alignment, not the locus of aggregation}, as the driver, and the
wrapper beats rotation-augmented training without retraining (\S\ref{sec:experiments}).
\item \textbf{An auditing procedure for rotation-natural domains.} A new way of benchmarking
explanation methods where the input has no canonical orientation, and the remedy that
follows: on histopathology and aerial imagery the wrapper is rotation-consistent
training-free (including zero-shot CLIP, $0.317\!\to\!0.777$), and its by-product
\textbf{PEUM} becomes a triage rule surfacing $4.4\times$ (aerial) and $2.8\times$
(histopathology) more unstable explanations than random at a $10\%$ budget
(\S\ref{sec:application}).
\item \textbf{Negative results others should know.} Score-CAM is numerically constant on
$81.4\%$ of VGG-16 inputs, XGrad-CAM is indistinguishable from Grad-CAM on ResNet-50
(agreeing to ${\sim}10^{-8}$), and ViT Grad-CAM works only when hooking \texttt{ln\_1}
(\S\ref{sec:experiments}). And about our own family: on a task-grounded, perturbation-free
check, \emph{no} CAM, ours included, concentrates much above a uniform map
(\S\ref{sec:application}).
\end{enumerate}

%% file: sec/2_related.tex
\section{Related Work}
\label{sec:related}

\noindent\textbf{Gradient-based CAM.}
CAM~\cite{cam} requires retraining; Grad-CAM~\cite{gradcam} removed that constraint with
gradient weights, and Grad-CAM++~\cite{gradcampp}, XGrad-CAM~\cite{xgradcam},
LayerCAM~\cite{layercam}, and Ablation-CAM~\cite{ablationcam} offer alternative weightings,
and recent work continues to refine \emph{what} a CAM explains (Finer-CAM~\cite{finercam}
and DiffCAM~\cite{diffcam} sharpen class discriminability) without changing that
each still conditions its heatmap on one orientation of the input. We also
observe two hidden pathologies (\S\ref{sec:experiments}): XGrad-CAM's per-channel
normalisation collapses to Grad-CAM's on ResNet-50's near-uniform $7{\times}7$ features
(maps identical to ${\sim}10^{-8}$, i.e.\ floating-point noise; \emph{exactly} identical on CUB IoU), and its axiom-based weights collapse to near-zero correlation
on ViT patch tokens.

\noindent\textbf{Perturbation- and optimisation-based.}
RISE~\cite{rise} and Score-CAM~\cite{scorecam} replace gradients with score-based weights;
Recipro-CAM~\cite{reciprocam} (gradient-free) and Opti-CAM~\cite{opticam}
(per-image weight optimisation) target gradient-noise sensitivity. Each remains
orientation-dependent, and Score-CAM's masking is numerically constant on $81.4\%$ of
VGG-16 inputs (\S\ref{sec:experiments}).

\noindent\textbf{Multi-view / augmentation-based, and why alignment matters.}
SmoothGrad~\cite{smoothgrad} and Smooth Grad-CAM++~\cite{smoothgradcampp} average over
input noise; Augmented Score-CAM~\cite{augscam} averages geometrically-augmented heatmaps.
These are the closest relatives of our method, and we position ourselves plainly:
\method{} is aligned test-time augmentation \emph{of the explanation}. The distinction that
matters is not novelty of ``averaging views'' but \emph{alignment}: inverse-rotating each
view's saliency to a common canonical frame before averaging. Averaging finished maps
\emph{without} this step (the route these methods take) can leave equivariance below the
single-view baseline on fragile backbones (\S\ref{sec:experiments}); alignment is what
turns view aggregation into an equivariant operator, and it applies to any pretrained CAM.

\noindent\textbf{Equivariant architectures.}
Group-equivariant CNNs~\cite{gcnn} build feature-level equivariance in through
architectural constraints, but require training (or retraining) the model. Our contribution
is complementary and \emph{post-hoc}: a training-free audit of the pretrained,
non-equivariant networks practitioners already deploy, exactly the setting where one
cannot retrain.

\noindent\textbf{Explanation invariance and equivariance.}
The demand that a saliency map track a symmetry is not ours to assume: Crabb\'{e} and van der
Schaar~\cite{crabbe2023robustness} argue that ``any explanation that faithfully explains'' a
$G$-invariant model ``needs to be in agreement with this invariance property,'' and show that
for a \emph{spatial} explanation the correct behaviour is equivariance; their score is, up to
the similarity function, our Eq (\S\ref{sec:setup}). Two things separate our work. In
\emph{scope}, they cover cyclic translations, permutations and the dihedral group on ECG,
FashionMNIST, point clouds and graphs, not continuous rotation, ImageNet-scale CNNs or ViTs,
or CAM-family explanations. In \emph{construction}, their Proposition~2.3 aggregates
$e(\rho[g]\mathbf{x})$ \emph{without} the inverse action, yielding an \emph{invariant}
explanation; for a spatial map that is exactly our \emph{unaligned} baseline, which
\S\ref{sec:experiments} shows sits at or \emph{below} single-view Grad-CAM on both CNNs. The
equivariant analogue, applying $\mathcal{T}_t^{-1}$ before averaging, is what works, and is
what \method{} does. Their Guideline~4 warrants enforcing consistency only when the model is
itself invariant; we do not assume this but \emph{measure} it (\S\ref{sec:signalnoise}) and
report Eq conditioned on the model's own stability, the approximately-invariant regime their
\S3.3 identifies as practical.

\noindent\textbf{Saliency consistency and uncertainty.}
Explanation fragility is well documented: saliency can be manipulated, is often
unreliable, and fails basic sanity checks~\cite{ghorbani2019fragile,kindermans2019unreliability,adebayo2018sanity};
COSE~\cite{cose} measures equivariance failures but does not remove them. Prior work thus
largely \emph{diagnoses} the problem or builds equivariance into the model. To our
knowledge \method{} is the first training-free, post-hoc method to make rotational
consistency a constructive objective for CAM-family explanations. U-CAM~\cite{ucam} and
BayesLIME~\cite{bayeslime} estimate explanation uncertainty but require model modification,
and the reliability of such estimates is itself under
scrutiny~\cite{sanityuncertainty,uqgrad}; PEUM (\S\ref{sec:application}) provides an
image-level signal for \emph{explanation} reliability, at no cost beyond the views the
wrapper has already taken.

%% file: sec/4_signalnoise.tex
\section{Is the Drift Signal or Noise?}
\label{sec:signalnoise}

Before removing the drift we must justify that removing it is faithful. The objection is
that a rotated input is partly out-of-distribution, so the network may genuinely compute
something different and its explanation \emph{should} change. A weaker version of the same
objection applies to us: a CAM is not computed from the pooled feature or the logits, so
showing that \emph{those} are rotation-stable does not establish that the quantities the
CAM actually reads are. We therefore open the operator and measure equivariance at every
stage of it, rather than inferring the answer from the network's output. Measurements are
on ImageNet-1K, $2{,}000$ images (one per class, $2$/class where noted) $\times$ $7$ angles.

\noindent\textbf{(1) Where in the operator does the drift enter?} Grad-CAM is a chain:
spatial activations $A$ and the class gradient field $g$ at the target layer, pooled into
channel weights $\alpha=\overline{g}$, combined as $\mathrm{ReLU}(\sum_k\alpha_k A_k)$, then
upsampled and normalised. Each stage is a spatial (or, for $\alpha$, a channel) object, so
each has its own equivariance. Table~\ref{tab:decomp} reports all of them. The last row
reproduces the published single-view Grad-CAM Eq of Table~\ref{tab:main} to within
$0.003$, which validates the decomposition against the deployed pipeline.

\begin{table}[t]
\centering\footnotesize
\caption{\textbf{Stage-wise equivariance of the Grad-CAM operator} (Eq$\uparrow$,
$n{=}2{,}000$, $7$ angles). The drift does \emph{not} enter at the channel-weighting step:
$\alpha$ is the most rotation-stable stage everywhere, and on ResNet-50 it is stable
\emph{exactly}. What moves is the spatial pair $(A,g)$. Yet the final map is far more
equivariant than its per-channel constituents, because summing over channels cancels much
of the per-channel error. $^{\ast}$ResNet-50's gradient field is spatially constant in every
one of its $2{,}048$ channels (std exactly $0$), so a spatial correlation is undefined; this
is a structural consequence of the GAP+linear head, not a measurement failure.}
\label{tab:decomp}
\setlength{\tabcolsep}{4pt}
\begin{tabular}{@{}lccc@{}}
\toprule
Stage of the operator & ResNet-50 & VGG-16 & ViT-B/16 \\
\midrule
Activations $A$                       & 0.321 & 0.246 & 0.453 \\
Gradient field $g$                    & ---$^{\ast}$ & 0.006 & 0.240 \\
Channel weights $\alpha$ (Pearson)     & \textbf{1.000} & \textbf{0.889} & \textbf{0.685} \\
\midrule
$\sum_k\alpha_k A_k$ (pre-ReLU)        & 0.708 & 0.492 & 0.265 \\
after ReLU                            & 0.708 & 0.500 & 0.240 \\
final map (= Grad-CAM)                & 0.700 & 0.511 & 0.247 \\
\emph{published Grad-CAM, Tab.~\ref{tab:main}} & \emph{0.703} & \emph{0.511} & \emph{0.250} \\
\bottomrule
\end{tabular}
\end{table}

\noindent The result is not the one we expected, and it is sharper. We had assumed the
instability entered where a single orientation is used to estimate the channel weights.
It does not: $\alpha$ is the \emph{most} stable stage on every backbone, and on ResNet-50
it is stable exactly, with cosine $1.000$ at every angle. The reason is structural. A
GAP+linear head makes $\partial y_c/\partial A_{kij}=W_{ck}/HW$, independent of $(i,j)$, so
$g$ is spatially constant in every channel and $\alpha$ depends only on the target class,
not on the orientation. We verified this directly: all $2{,}048$ channels have spatial
standard deviation exactly zero. Consequently \emph{every bit} of ResNet-50's Grad-CAM
rotation drift is carried by the activation tensor.

\noindent\textbf{(2) The model discards what Grad-CAM varies over.} This localisation is
what licenses the paper's claim, and it replaces the weaker argument from output
similarity. The penultimate feature and logit vectors of $\mathbf{x}$ and
$R_\theta(\mathbf{x})$ are indeed similar, cosine $0.90$ / $0.85$ over all $14{,}000$
image--angle pairs, and not trivially so: between \emph{unrelated} images the same cosines
are $0.529$ / $0.088$. But Table~\ref{tab:decomp} shows \emph{why} they are high, and it is
not that the spatial computation is unchanged, which it plainly is not ($0.321$ on
ResNet-50). It is that global pooling discards exactly the spatial arrangement that
rotation disturbs. For a GAP+linear classifier this is exact rather than suggestive: the
decision is a function of $\mathrm{GAP}(A)$ alone, so any rotation-induced rearrangement
\emph{within} $A$ that leaves the pooled vector fixed is, by construction, invisible to the
model and fully visible to Grad-CAM. The drift is variation along directions the classifier
provably throws away. That is a precise sense in which it is operator noise, and it is the
sense we claim.

\noindent\textbf{(3) The drift is largely prediction-independent.} On the $64.3\%$ of
rotations leaving the top-1 unchanged (where the model has \emph{not} changed its decision),
Grad-CAM equivariance is only $0.758$ while \method{} reaches $0.959$. Per-image drift
($1-\text{Eq}$) there is only weakly correlated with the change in confidence
($r{=}0.060$, $p{=}2.9\!\times\!10^{-8}$, $n{=}8{,}646$ stable pairs): a small faithful
component is detectable at this sample size but explains $0.4\%$ of the variance, so the
drift is dominated by a prediction-independent floor.

\begin{figure}[t]
\centering
\includegraphics[width=0.85\linewidth]{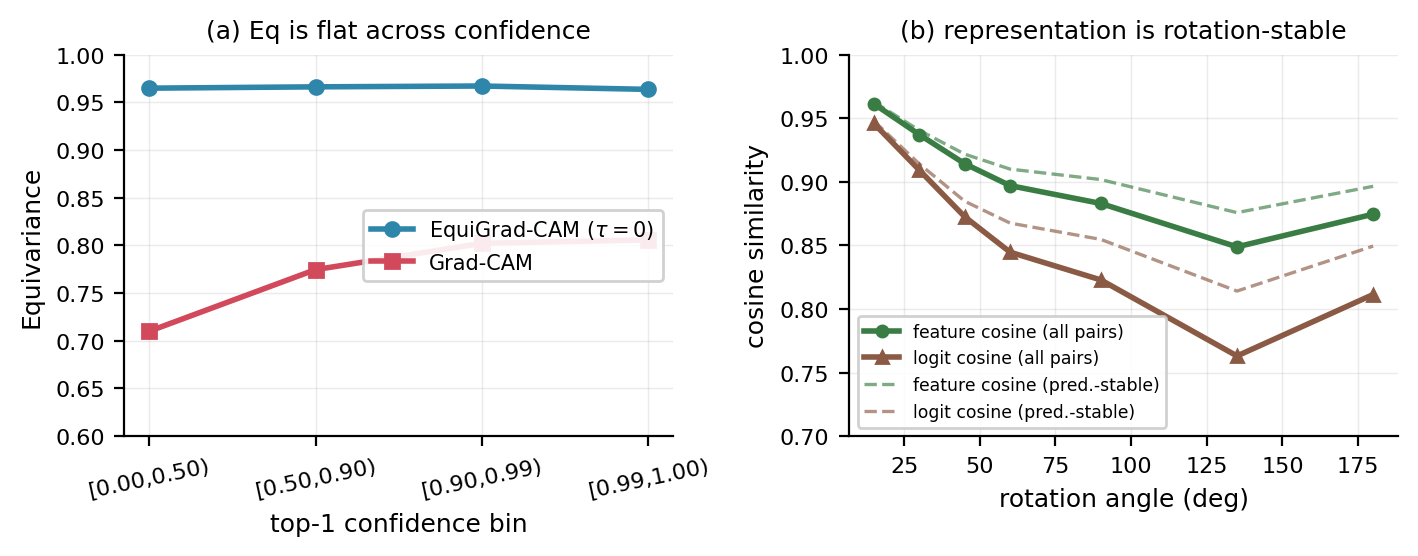}
\caption{\textbf{The drift does not track the model's response.} (a) \method{}'s
equivariance is flat across confidence strata while single-view Grad-CAM's stays far lower
even at high confidence, so the drift is not confined to inputs the model finds hard.
(b) the network's \emph{pooled} feature/logit representation is nearly rotation-invariant.
Table~\ref{tab:decomp} shows why this is not the same as the spatial computation being
unchanged: global pooling discards precisely the spatial arrangement rotation disturbs,
which is the sense in which Grad-CAM's drift is operator noise.}
\label{fig:signalnoise}
\end{figure}

\noindent\textbf{(4) Even the most confident, stable inputs drift.} Stratifying by top-1
probability, \method{}'s equivariance is flat ($\approx0.965$) across all bins whereas
Grad-CAM's climbs only $0.71\!\to\!0.81$ and never approaches consistency, even on the
highest-confidence inputs (Fig.~\ref{fig:signalnoise}). A confidence-weighted equivariance,
down-weighting views with large confidence drop exactly as the objection requests, leaves the
ranking unchanged ($0.739$ vs.\ $0.959$).

\noindent\textbf{(5) A causal test: the drift lives where the model does not look.} The
measurements so far are correlational. Here is a direct one. For each (image, angle) we
define three regions \emph{in the canonical frame} and then transport them to the rotated
frame, so the same pixels are tested twice: the \emph{drift} region (top $10\%$ of
$|h_0-\tilde h_\theta|$), the \emph{agreement} region (top $10\%$ of
$\min(h_0,\tilde h_\theta)$), and a \emph{random} region of equal area. We occlude each with
the Gaussian-blur baseline already used for insertion/deletion ($\sigma{=}10$) and record
the drop in the model's probability for its original top-1. If the drift carried
orientation-specific evidence, occluding it should matter at one orientation and not the
other.

\noindent It does not, and it barely matters at either. On ResNet-50 ($n{=}1{,}000$, one
image per class) the drift region is the \emph{least} consequential of the three: blurring
it costs $0.100$ in the canonical frame and $0.038$ in the rotated one, against
$0.152$/$0.121$ for a random region and $0.264$/$0.225$ for the agreement region. The
pixels whose saliency moves under rotation are pixels the model relies on \emph{less than
pixels chosen at random}, and $2.6\times$ less than the pixels where the explanation is
stable. The orientation-asymmetry test agrees: mean $|\Delta_0-\Delta_\theta|$ is $0.146$
$[0.142,0.151]$ for the drift region against $0.210$ $[0.205,0.215]$ for random, so the
faithful-signal direction fails outright (one-sided Wilcoxon, $p{=}1.00$, median difference
${-}0.021$). VGG-16 returns the same null ($0.218$ $[0.213,0.223]$ vs.\ $0.220$
$[0.215,0.224]$, $p{=}0.96$).

\noindent We rest the argument on the first of those results rather than the second, for a
reason worth stating: the drift region is defined by a \emph{disagreement} between the two
frames, so it is asymmetric by construction and the asymmetry test is tilted toward finding
faithful signal. That nothing of consequence appears even under a tilted test is
informative, but the load-bearing observation is the below-random causal importance, which
no construction artefact explains.

\noindent\textbf{Reporting equivariance conditioned on the model.} The premise we are
testing, that a faithful explanation of a rotation-stable model should itself be
rotation-equivariant, is licensed only to the extent the model \emph{is} stable, a
condition made explicit by Crabb\'{e} and van der Schaar~\cite{crabbe2023robustness}
(their Guideline~4). We therefore do not report a single unconditional number and leave the
condition implicit. Alongside the unconditional $\text{Eq}$ used in the tables (for
comparability with prior work), we report it restricted to the $64.3\%$ of rotations that
leave the top-1 unchanged (where the model demonstrably did not change its decision, so
equivariance is unambiguously the right target), and a continuously
confidence-weighted variant that relaxes the requirement in proportion to how far the
model's response moved, rather than gating it discretely. Each triple below is computed on the same
$2{,}000$ images, so the three numbers differ only in conditioning and not in sample
(unconditional / prediction-stable / confidence-weighted):

\vspace{2pt}
\noindent\hspace*{\fill}\begin{tabular}{@{}lcc@{}}
\toprule
Backbone & Grad-CAM & \method{} \\
\midrule
ResNet-50 & 0.711 / 0.758 / 0.739 & 0.958 / 0.959 / 0.959 \\
VGG-16    & 0.513 / 0.580 / 0.562 & 0.958 / 0.963 / 0.963 \\
ViT-B/16  & 0.259 / 0.284 / 0.276 & 0.867 / 0.850 / 0.856 \\
\bottomrule
\end{tabular}\hspace*{\fill}
\vspace{3pt}

\noindent Conditioning on the model's own stability moves Grad-CAM by at most $+0.067$
(VGG-16, whose top-1 survives rotation least often at $56.9\%$) and \method{} by less than
$0.017$ in either direction. It never closes a gap that runs from $0.25$ to $0.61$. The
conclusion is therefore insensitive to which conditioning one accepts, which is why we
report all three rather than pick the one that flatters us.

\noindent\textbf{Equivariance is not bought by smoothing.} A high Eq could in principle be
obtained by returning a blurrier, less discriminative map rather than a better-aligned one.
It cannot: Gaussian-blurring single-view Grad-CAM's own map at $\sigma\in\{2,4,8,16\}$ leaves
its equivariance flat at $0.700$--$0.702$, against $0.700$ unblurred ($n{=}2{,}000$). The
metric does not reward low-frequency maps, so the gap in Table~\ref{tab:align} cannot be
explained by our aggregate being smoother.

\noindent\textbf{Conclusion, stated at the strength the evidence supports.} We do not
claim the network computes the same thing when the image is turned; Table~\ref{tab:decomp}
shows it does not, and the spatial activations move a great deal. We claim something
narrower and better supported: \emph{the drift Grad-CAM reports is carried by degrees of
freedom the classifier discards}. On ResNet-50 that is exact, because the head reads only
$\mathrm{GAP}(A)$ and $\alpha$ is orientation-invariant by construction; on VGG-16
and ViT-B/16 it is the same picture with the pooling argument weakened to an empirical one.
Aligning and averaging views (\S\ref{sec:method}) therefore recovers the explanation the
model's own decision already implies, rather than erasing a faithful change. It remains
\emph{mostly}, not always: on the $35.7\%$ of rotations that flip the top-1 the model does
respond differently, and there the honest operating point is the prediction-conditioned
\methodpcf{}, while PEUM flags the least reproducible explanations for human review
(\S\ref{sec:application}). The objection thus becomes a motivation for our own gate and
triage, not a barrier to aggregation.

%% file: sec/3_method.tex
\section{Method}
\label{sec:method}

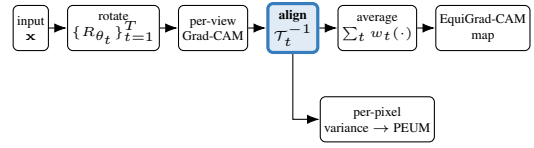
\begin{figure}[t]
\centering
\begin{tikzpicture}[
  font=\tiny,
  box/.style={draw, rounded corners=2pt, minimum height=6mm, align=center, inner sep=1.5pt},
  algn/.style={draw, rounded corners=2pt, minimum height=6.5mm, align=center, inner sep=2pt,
               fill=wacvblue!15, very thick, draw=wacvblue},
  >=Latex, node distance=2.4mm and 2.4mm]
\node[box] (x) {input\\$\mathbf{x}$};
\node[box, right=of x] (rot) {rotate\\$\{R_{\theta_t}\}_{t=1}^{T}$};
\node[box, right=of rot] (gc) {per-view\\Grad-CAM};
\node[algn, right=of gc] (align) {\textbf{align}\\$\mathcal{T}_t^{-1}$};
\node[box, right=of align] (avg) {average\\$\sum_t w_t(\cdot)$};
\node[box, right=of avg] (out) {\method{}\\map};
\node[box, below=6mm of avg] (peum) {per-pixel\\variance $\to$ PEUM};
\draw[->] (x) -- (rot); \draw[->] (rot) -- (gc); \draw[->] (gc) -- (align);
\draw[->] (align) -- (avg); \draw[->] (avg) -- (out);
\draw[->] (align) |- (peum);
\end{tikzpicture}
\caption{\textbf{\method{} pipeline.} $T$ rotated views each get a Grad-CAM pass; each
view's saliency is inverse-rotated back to the canonical frame (\textbf{align}) and the
aligned views are averaged. ``Canonical frame'' means the \emph{computational} reference
frame of the presented input, not a claim that the content has a correct orientation---in
these domains it does not, which is why any single presented orientation is an arbitrary
basis for an explanation. The aligned views' per-pixel variance is PEUM, at no cost beyond
the views already taken. Align is what makes the average equivariant; without it, averaging
finished maps can fall \emph{below} single-view Grad-CAM (\S\ref{sec:experiments}).}
\label{fig:pipeline}
\end{figure}

\noindent\textbf{Preliminaries.} Let $A^k\in\mathbb{R}^{h\times w}$ be channel $k$'s
activation at the target layer, $g^k=\partial y^c/\partial A^k$ its gradient for class-$c$
logit $y^c$. Grad-CAM~\cite{gradcam} is
$L_{\text{GC}}=\mathrm{ReLU}\!\left(\sum_k\alpha_k A^k\right)\uparrow_{H\times W}$ with
$\alpha_k=\frac{1}{hw}\sum_{i,j}g^k_{ij}$; all maps are min--max normalised to $[0,1]$.

\subsection{Aligned Multi-View Aggregation}

Given image $\mathbf{x}$ and target class $c$ (the model's top-1 on the unrotated image,
held fixed across views), we sample $T{=}18$ angles on a uniform grid and form views
$\mathbf{x}_t=R_{\theta_t}(\mathbf{x})$, each passed forward--backward through the frozen
backbone (Fig.~\ref{fig:pipeline}). The operative step is \textbf{alignment}: every per-view
quantity is inverse-rotated to the canonical frame, $\mathcal{T}_t^{-1}(\cdot)$, at native
resolution $h{\times}w$ (bilinear, $64$-px reflective padding) \emph{before} averaging. Only
then do consistent explanations reinforce and view-specific operator noise cancel.

We instantiate the aligned average at two loci, near-identical at full scale
(\S\ref{sec:experiments}), so \emph{alignment, not the locus, is what matters}.
\textbf{(i) Feature space} (headline): average aligned activations and gradients,
$\bar{A}^k=\sum_t w_t\,\mathcal{T}_t^{-1}(A^k_t)$,
$\bar{g}^k=\sum_t w_t\,\mathcal{T}_t^{-1}(g^k_t)$, then form one map,
\begin{equation}
L_{\text{EG}}=\mathrm{ReLU}\!\left(\textstyle\sum_k \bar{\alpha}_k \bar{A}^k\right)\uparrow_{H\times W},
\quad \bar{\alpha}_k=\tfrac{1}{hw}\textstyle\sum_{i,j}\bar{g}^k_{ij}.
\label{eq:agg}
\end{equation}
This averages gradients \emph{before} the ReLU (enabling pre-nonlinearity denoising) and
upsamples once (avoiding $T$-fold grid artefacts). \textbf{(ii) Aligned output space}:
average the finished per-view maps, $L^{\text{out}}=\sum_t w_t\,\mathcal{T}_t^{-1}(L_{\text{GC}}(\mathbf{x}_t))$.
The contrast that isolates our claim is with \emph{unaligned} output-space averaging
($\sum_t w_t L_{\text{GC}}(\mathbf{x}_t)$, no $\mathcal{T}_t^{-1}$), the route of prior
multi-view CAMs, which is not equivariant (\S\ref{sec:experiments}).

\subsection{A Family of Operating Points}
\label{sec:tau}

The weights $w_t$ index a one-parameter family. \textbf{Headline (\method{}, $\tau{=}0$):}
uniform $w_t{=}1/T$; every view contributes, giving maximum noise reduction and the highest
equivariance. \textbf{Prediction-conditioned (\methodpcf{}, $\tau{=}0.10$):} Predictive
Confidence Filtering accepts view $t$ iff $\hat{y}(\mathbf{x}_t){=}c$ and
$\hat{p}(c|\mathbf{x}_t){\geq}\tau$ and weights accepted views by confidence, so views whose
top-1 flips are excluded rather than smoothed in. The headline uses \emph{no} gate, so the
binary threshold is not load-bearing: a continuous soft gate
($w_t\propto\hat{p}(c|\mathbf{x}_t)^{\gamma}$) reaches Eq $0.962$--$0.966$ against the ungated
$0.966$ ($n{=}2{,}000$). One implementation spans both points.

\subsection{ViT Extension and PEUM}

\noindent\textbf{ViT.} For ViT-B/16 we reshape the $196$ patch tokens to $14{\times}14$ and
hook \texttt{encoder.layers[-1].ln\_1}. This location is critical: \texttt{ln\_2} or the full
block yields $100\%$ constant heatmaps, because the residual addition distributes gradients
uniformly across patches (\S\ref{sec:experiments}). We therefore claim
architecture-agnosticism for the \emph{aggregation}, which needs only a spatial map and its
inverse action, not for the CAM extraction, which on a transformer still requires a hook that
produces a non-degenerate map.

\noindent\textbf{PEUM.} The aligned per-view maps $\{h_t\}$ needed for Eq.~\ref{eq:agg} also
give, at no cost beyond the aggregation, a per-pixel weighted variance
\begin{equation}
\sigma^2_{ij}=\textstyle\sum_{t} w_t\bigl(h_t(i,j)-\bar{h}(i,j)\bigr)^2,
\label{eq:peum}
\end{equation}
whose image-level mean $\bar\sigma^2=\frac{1}{HW}\sum_{ij}\sigma^2_{ij}$ ranks images by
how reproducible their explanation is under rotation: a high value flags a map an auditor
should not trust without looking. We validate PEUM as an \emph{image-level} signal
(per-pixel calibration is weaker), and show in \S\ref{sec:application} that it ranks
\emph{explanation} instability better than any cheaper signal we tested while \emph{not}
predicting model error, which is a different quantity and better served by the
prediction-flip rate.

\subsection{Evaluation Protocol}
\label{sec:setup}

\noindent Equivariance is
$\text{Eq}=\rho\!\left(h(R_\theta\mathbf{x}),R_\theta h(\mathbf{x})\right)$, the Pearson
correlation between the heatmap of the rotated image and the rotated heatmap of the original,
averaged over $\{15,30,45,60,90,135,180\}^\circ$; five lie off the internal $T{=}18$ grid, so
the gains are not an artefact of probing aggregated angles. Faithfulness is insertion (higher
better) / deletion (lower better) AUC over $20$ steps against a Gaussian-blur baseline
($\sigma{=}10$), reducing the missingness bias of zero-masking~\cite{jain2022missingness,balasubramanian2022improved,rong2022consistent,agarwal2020removing}.
Significance is a paired one-sided Wilcoxon signed-rank test with Bonferroni correction;
\texttt{***} denotes $p{<}10^{-200}$ (paired $p$ saturate the \texttt{float64} floor at our
$n$). Pearson is undefined for a constant heatmap ($\sigma{<}10^{-8}$); we assign
$\text{Eq}{=}0$ (the worst case) and apply it to \emph{every} method including our own, since
a constant map has explained nothing. It is material only for Score-CAM, constant on $17.6\%$
(R50) and $81.4\%$ (VGG) of inputs; the largest fraction for any other method is $3.0\%$ (our
own aggregate on zero-shot CLIP) and $\leq0.2\%$ on ImageNet. Insertion and deletion are
defined for constant maps and use every image.

%% file: sec/6_application.tex
\section{Application: Auditing in Rotation-Natural Domains}
\label{sec:application}

The equivariance defect matters most where the input has no canonical orientation, because
there two rotations are equally valid views and a drifting explanation cannot serve as
evidence. We study two such domains as a saliency-auditing application: \textbf{digital
histopathology} (PatchCamelyon~\cite{pcam}, H\&E tumour patches, the benchmark
introduced to study rotation equivariance in pathology, where a slide is rotated freely)
and \textbf{aerial scene classification} (RESISC45~\cite{resisc45}, overhead tiles with no
intrinsic ``up'').

\noindent\textbf{Protocol.} We fine-tune an ImageNet ResNet-50 on each domain with a standard
recipe and, deliberately, \emph{no rotation augmentation}, keeping the model rotation-naive,
so any equivariance the \emph{explanation} gains is attributable to the wrapper, not the
weights. The classifiers reach $82.2\%$ (PatchCamelyon; a train/test hospital shift makes
this realistic and non-saturated) and $95.2\%$ (RESISC45). We then run the
evaluation protocol of \S\ref{sec:setup} (Eq / Ins / Del over the same seven angles, plus
PEUM) on $1{,}000$ test
images per domain. Because we trained these two models ourselves, unlike the pretrained
backbones of \S\ref{sec:experiments} where seeds do not apply, we repeat each fine-tune
over three seeds. Clean accuracy varies by $\pm1.3$ points (histopathology) and $\pm0.1$
(aerial), and the gap of Table~\ref{tab:application} is unmoved: Grad-CAM
$0.489{\pm}0.016$ vs.\ \method{} $0.925{\pm}0.005$ on histopathology, and
$0.717{\pm}0.004$ vs.\ $0.953{\pm}0.001$ on aerial.

\begin{table}[t]
\centering\footnotesize
\caption{\textbf{Auditing two rotation-natural domains} ($n{=}1{,}000$). \method{} restores
equivariant explanations training-free on both. On histopathology single-view Grad-CAM is
worst of any domain ($0.484$), and \emph{unaligned} averaging is worse still ($0.412$,
below Grad-CAM), the cleanest evidence that alignment, not averaging, is the mechanism.
Faithfulness (Ins/Del) reported descriptively. $^{\ddagger}$Aug.\ Score-CAM is Eq-only
(its inner Score-CAM costs one forward pass per channel, per view) and is the strongest
multi-view competitor; it is scored under the same uniform $\text{Eq}{=}0$ convention as
every other row, and is numerically constant on $9.6\%$ (histopathology) and $28.3\%$
(aerial) of images, on which its conditional means are $0.909$ and $0.879$. Repeating the
fine-tune over three seeds moves \method{} by $\pm0.005$ (histopathology) and $\pm0.001$
(aerial). PEUM image-level triage: $r{=}0.52$ (histopathology), $r{=}0.68$ (aerial).}
\label{tab:application}
\begin{tabular}{@{}lccc@{}}
\toprule
Method & Eq$\uparrow$ & Ins$\uparrow$ & Del$\downarrow$ \\
\midrule
\multicolumn{4}{@{}l}{\emph{Histopathology --- PatchCamelyon} (test acc.\ $82.2\%$)} \\
Grad-CAM                        & 0.484 & 0.892 & 0.650 \\
Unaligned averaging             & 0.412 & 0.839 & 0.743 \\
Aug.\ Score-CAM$^{\ddagger}$     & 0.822 & --- & --- \\
\methodpcf{} ($\tau{=}0.10$)    & 0.921 & 0.875 & 0.687 \\
\textbf{\method{}} ($\tau{=}0$) & \textbf{0.924} & 0.874 & 0.689 \\
\midrule
\multicolumn{4}{@{}l}{\emph{Aerial --- RESISC45} (test acc.\ $95.2\%$)} \\
Grad-CAM                        & 0.713 & 0.616 & 0.373 \\
Unaligned averaging             & 0.729 & 0.498 & 0.434 \\
Aug.\ Score-CAM$^{\ddagger}$     & 0.630 & --- & --- \\
\methodpcf{} ($\tau{=}0.10$)    & 0.949 & 0.587 & 0.376 \\
\textbf{\method{}} ($\tau{=}0$) & \textbf{0.952} & 0.587 & 0.376 \\
\bottomrule
\end{tabular}
\end{table}

\begin{figure}[t]
\centering
\includegraphics[width=0.85\linewidth]{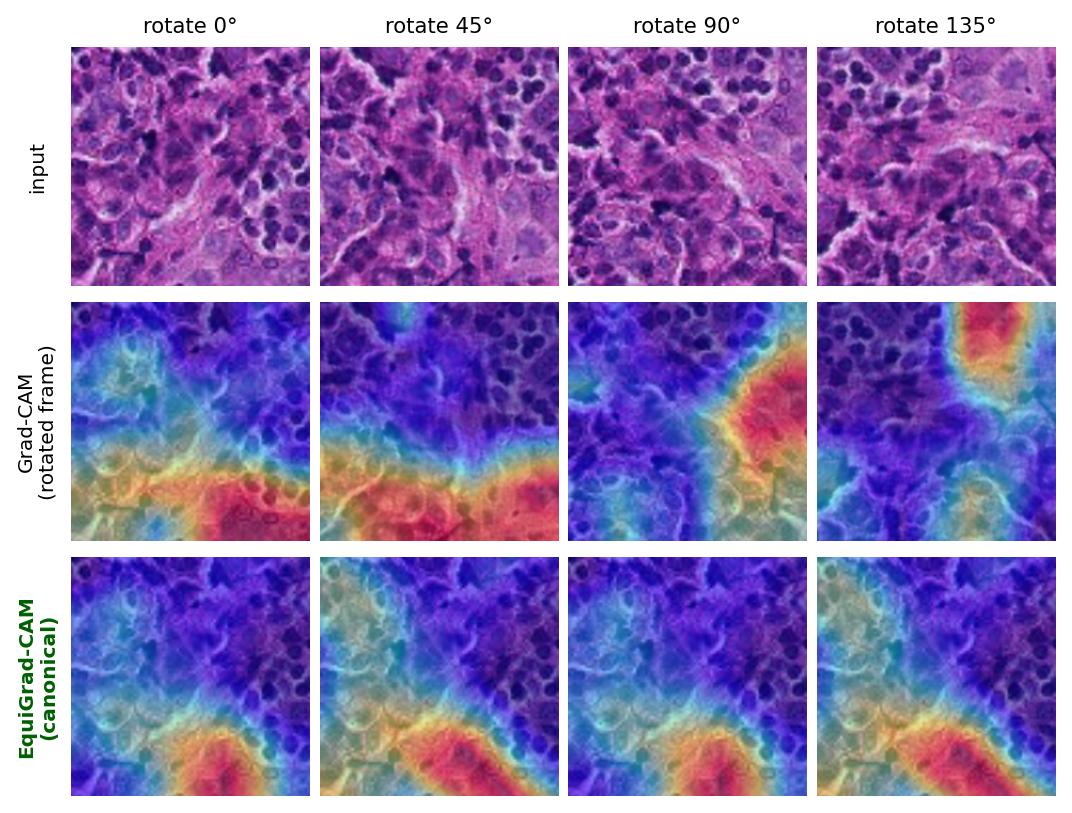}
\caption{\textbf{Explanation drift under rotation, histopathology.} Each map is computed on
the \emph{rotated} input. \emph{Top:} Grad-CAM, shown in the rotated frame, which drifts.
\emph{Bottom:} \method{}, inverse-rotated to the canonical frame, where an equivariant method
yields the same map across the row.}
\label{fig:eqdemo}
\end{figure}

\noindent\textbf{Findings.} Table~\ref{tab:application} tells the same story in both domains,
most sharply in histopathology (Fig.~\ref{fig:eqdemo}). (i) Single-view Grad-CAM is
\emph{least} equivariant here of any \emph{domain} we tested at ResNet-50 ($0.484$): with no
canonical orientation its map drifts most. (ii) \method{} makes the explanation
rotation-consistent training-free, $0.484\!\to\!0.924$ ($+90.9\%$) and $0.713\!\to\!0.952$
($+33.5\%$), on models never trained for rotation. (iii) \emph{Unaligned} averaging reaches
only $0.412$ on histopathology, \emph{below} Grad-CAM: absent alignment, pooling views hurts.
As on the ImageNet CNNs, this carries a small faithfulness cost (PCam insertion
$0.892\!\to\!0.874$, deletion $0.650\!\to\!0.689$, both $p{<}10^{-14}$; RESISC insertion
$0.616\!\to\!0.587$, $p{<}10^{-25}$, deletion n.s.), reported descriptively.

\noindent\textbf{Would training for rotation have been enough?} The obvious alternative to
a post-hoc wrapper is to train the model to be rotation-invariant, so we re-trained both
domain models with rotation augmentation (uniform in $[-180^\circ,180^\circ]$, applied with
the \emph{same} operator the evaluation uses) and re-ran the protocol. It helps the
single-view baseline and does not substitute for alignment: Grad-CAM equivariance rises
$0.484\!\to\!0.617$ on histopathology and $0.713\!\to\!0.752$ on aerial, while clean accuracy
\emph{falls} $82.2\!\to\!80.8\%$ and $95.2\!\to\!94.1\%$. Both augmented models remain far below
the wrapper applied to the un-augmented model ($0.924$ / $0.952$), and \method{} on the
augmented model is higher still ($0.935$ / $0.957$). Equivariant training and equivariant
explanation are complementary, and only the latter is available when auditing a model one
cannot retrain.

\noindent\textbf{Does the explanation point at the tissue that defines the label?}
Insertion/deletion are perturbation metrics, and perturbation is exactly what the
missingness-bias literature warns
about~\cite{jain2022missingness,balasubramanian2022improved,rong2022consistent,agarwal2020removing}. PatchCamelyon
lets us sidestep that entirely, because its label is defined by \emph{geometry}: a patch is
positive iff the centre $32{\times}32$ region of the $96{\times}96$ patch contains tumour
tissue, and tumour in the outer margin does not affect the label~\cite{pcam}. We therefore
know, with no annotation, which part of a positive patch the label is about. Under the
frozen Resize(256)+CenterCrop(224) transform that region becomes a centred $85.3$\,px box;
we score against a centred disc of equal area, which is \emph{exactly} invariant to rotation
about the image centre, and measure the fraction of saliency mass inside it. A spatially
uniform map scores the disc's area fraction, $0.145$.

\begin{table}[t]
\centering\footnotesize
\setlength{\tabcolsep}{4pt}
\caption{\textbf{Label-region energy on PatchCamelyon positives} ($n{\approx}460$). Fraction
of saliency mass inside the label-defining centre region, at $0^\circ$ and averaged over the
seven rotations. \emph{No} CAM concentrates strongly on the region that defines the label
(uniform-map null $0.145$), and ours is mid-pack on the level; what distinguishes the
aligned aggregates is how little they \emph{lose} when the input rotates.}
\label{tab:labelregion}
\begin{tabular}{@{}lccc@{}}
\toprule
Method & $0^\circ$ & rotated & drop$\downarrow$ \\
\midrule
\emph{uniform map (null)} & \emph{0.145} & \emph{0.145} & \emph{0.000} \\
Grad-CAM                        & 0.167 & 0.161 & 0.0060 \\
Grad-CAM++                      & \textbf{0.185} & 0.177 & 0.0080 \\
LayerCAM                        & \textbf{0.185} & 0.177 & 0.0080 \\
AugCAM ($\tau{=}0$, unaligned)  & 0.180 & 0.177 & 0.0028 \\
\method{} ($\tau{=}0$)          & 0.179 & 0.174 & 0.0048 \\
\methodpcf{} ($\tau{=}0.10$)    & 0.183 & 0.181 & \textbf{0.0019} \\
\bottomrule
\end{tabular}
\end{table}

\noindent Table~\ref{tab:labelregion} reports a result we did not expect. \emph{No} CAM
concentrates much mass on the label-defining region: every method lands between $0.167$ and
$0.185$ against a null of $0.145$. On this task-grounded, perturbation-free measure the CAM
family as a whole is only weakly localised on the evidence the label derives from, a caution
no insertion/deletion number reveals. Ours is \emph{not} best on the level ($0.179$;
Grad-CAM++ and LayerCAM reach $0.185$). What the aligned aggregates win is \emph{stability}:
\methodpcf{} loses only $0.0019$ of its label-region energy under rotation against $0.0060$
for Grad-CAM and $0.0080$ for Grad-CAM++/LayerCAM. Alignment keeps an explanation pointing at
the same tissue when the slide is turned; it does not make that tissue the right tissue.

\noindent\textbf{PEUM as a triage rule.} An auditor cannot inspect every case, so the
operational question is not whether PEUM correlates with instability. It does, at
$r{=}0.68$ (aerial) and $r{=}0.52$ (histopathology), both $p{<}10^{-69}$
(supplement). The question is how much a \emph{fixed review budget} buys. Ranking images by
PEUM and inspecting the top $10\%$ surfaces $44.4\%$ of the worst-decile (least
rotation-stable) explanations on aerial and $28.0\%$ on histopathology, against $10\%$ for
random triage: a $4.4\times$ and $2.8\times$ lift. At a $20\%$ budget the figures are
$68.7\%$ and $51.0\%$. These are exactly the cases, including the prediction-flipping
rotations of \S\ref{sec:signalnoise}, that a human should re-check, and the ranking costs
nothing beyond the views \method{} has already computed.

\noindent\textbf{What PEUM does, and does not, predict.} Two checks on
ImageNet/ResNet-50 ($n{=}2{,}000$, one image per class) sharpen this claim and bound it.
First, the operating point is not fitted to the images it is scored on: choosing the
top-decile threshold on one half of the sample and measuring lift on the held-out half
gives $1.97\times$, against $1.97\times$ in-sample, so the ranking is not overfitted.
Among the signals an auditor already has for free, PEUM is the strongest predictor of
explanation instability ($r{=}0.83$, versus $0.73$ for pairwise disagreement among the
aligned views, $0.24$ for prediction-flip rate and $0.20$ for confidence drop; paired
bootstrap on PEUM minus CAM-disagreement $={+}0.10$, $95\%$ CI $[0.08,0.12]$).
Second, and \emph{against} our own expectation, PEUM does not predict whether the model
is \emph{wrong}: its AUROC for misclassification is $0.578$ $[0.549,0.606]$, below both
prediction-flip rate ($0.764$) and confidence drop ($0.679$), which are cheaper still. We
therefore scope PEUM deliberately: it ranks explanations by how reproducible they are
under rotation, not predictions by how likely they are to be wrong. An auditor who wants
the latter should rank by flip rate; the two signals are complementary, and conflating
them would overstate what a per-pixel variance map can deliver.

\noindent\textbf{Does the audit hold for a foundation model?} The CAM literature is written
around supervised CNNs, so we repeat the aerial audit on \textbf{CLIP ViT-B/16 zero-shot}:
the classifier head is the text encoder's class-prompt embeddings, with no fine-tuning, so
this is the artefact a practitioner would deploy. It reaches $63.1\%$ top-1 on RESISC45, and
we explain its own top-1 with the same protocol and angles. Grad-CAM's equivariance is
$0.317$, as fragile as on the supervised ViT, and \method{} raises it to $\mathbf{0.777}$
($+145\%$), the largest relative gain we measure anywhere. This is after charging our own
method $\text{Eq}{=}0$ on the $3.0\%$ of images where its aggregate collapses to a constant
map, a failure mode single-view Grad-CAM does not have here and which we report rather than
condition away. As on the supervised ViT, the gain is not paid for in faithfulness:
insertion \emph{improves} $0.309\!\to\!0.334$ and deletion $0.223\!\to\!0.222$.

\begin{table}[t]
\centering\footnotesize
\setlength{\tabcolsep}{5pt}
\caption{\textbf{CLIP ViT-B/16, zero-shot on RESISC45} ($n{=}1{,}000$). The equivariance
defect is not an artefact of supervised training, and on this backbone removing it costs
nothing on Ins/Del. Eq applies the constant-map convention of \S\ref{sec:setup} uniformly:
our aggregate is numerically constant on $3.0\%$ of these images (the largest such rate for
any non-Score-CAM cell in the paper) and scores $0$ there; on the $970$ where it is defined
it averages $0.801$.}
\label{tab:clip}
\begin{tabular}{@{}lccc@{}}
\toprule
Method & Eq$\uparrow$ & Ins$\uparrow$ & Del$\downarrow$ \\
\midrule
Grad-CAM                        & 0.317 & 0.309 & 0.223 \\
Grad-CAM++                      & 0.137 & 0.288 & 0.243 \\
LayerCAM                        & 0.205 & 0.233 & 0.286 \\
AugCAM ($\tau{=}0$, unaligned)  & 0.385 & 0.306 & 0.240 \\
\methodpcf{} ($\tau{=}0.10$)    & 0.712 & 0.333 & 0.226 \\
\textbf{\method{}} ($\tau{=}0$) & \textbf{0.777} & \textbf{0.334} & \textbf{0.222} \\
\bottomrule
\end{tabular}
\end{table}

%% file: sec/5_imagenet.tex
\section{Generality: ImageNet-Scale Validation}
\label{sec:experiments}

\noindent\textbf{Setup.} Having established the audit and the remedy in the two
rotation-natural domains, we now test whether the same picture holds at ImageNet scale and
across architectures. We evaluate on the full official ILSVRC-2012 ImageNet-1K
validation set~\cite{deng2009imagenet} ($10{,}000$ images, $10$/class over all $1{,}000$
classes, $224{\times}224$ after Resize(256)+CenterCrop(224)) and CUB-200-2011~\cite{cub}.
Backbones are ImageNet-pretrained ResNet-50~\cite{resnet}
($7{\times}7$), VGG-16~\cite{vgg} ($14{\times}14$), and ViT-B/16~\cite{vit}. Metrics and
significance testing are as defined in \S\ref{sec:setup}.

\begin{table}[t]
\centering\footnotesize
\caption{\textbf{Rotational equivariance on ImageNet-1K} (Eq$\uparrow$).
\method{} ($\tau{=}0$) is highest on every backbone. $n$ is the number of images scored
per cell; rows marked $^{\dagger}$ are compute-bound and run at reduced $n$
(\S\ref{sec:matched}), which we account for explicitly rather than compare across
different samples. Score-CAM and Aug.\ Score-CAM weight \emph{channels} by masked-input
scores and have no channel axis to weight on patch tokens ($-$). RISE has no such
obstacle---it is black-box input masking and never touches tokens---so we report it on
ViT as well. Faithfulness is discussed in the text.}
\label{tab:main}
\setlength{\tabcolsep}{3.5pt}
\begin{tabular}{@{}lcccc@{}}
\toprule
Method & R50 & VGG & ViT & $n$ \\
\midrule
Grad-CAM             & 0.703 & 0.511 & 0.250 & 10k \\
Grad-CAM++           & 0.718 & 0.580 & 0.181 & 10k \\
LayerCAM             & 0.718 & 0.439 & 0.248 & 10k \\
Score-CAM            & 0.494 & 0.102 & --- & 10k \\
RISE$^{\dagger}$     & 0.421 & 0.384 & 0.306 & 2k/300 \\
Aug.\ Score-CAM$^{\dagger}$ & 0.907 & 0.766 & --- & 979/697 \\
Recipro-CAM$^{\dagger}$ & 0.763 & 0.631 & 0.191 & 2k \\
AugCAM ($\tau{=}0$, unaligned) & 0.690 & 0.495 & 0.317 & 10k \\
\methodpcf{} ($\tau{=}0.10$) & 0.926 & 0.886 & 0.768 & 10k \\
\textbf{\method{}} ($\tau{=}0$) & \textbf{0.956} & \textbf{0.958} & \textbf{0.867} & 10k \\
\bottomrule
\end{tabular}
\end{table}

\noindent\textbf{Equivariance.} Table~\ref{tab:main} gives the headline. At $\tau{=}0$,
equivariance over Grad-CAM rises $0.703\!\to\!0.956$ ($+36.0\%$) on ResNet-50,
$0.511\!\to\!0.958$ ($+87.5\%$) on VGG-16, and $0.250\!\to\!0.867$ ($+247\%$) on ViT-B/16;
every comparison is significant after correction. No baseline is close, and the margin is
if anything understated by Table~\ref{tab:main}: read on a \emph{matched} sample
(\S\ref{sec:matched}) the strongest multi-view competitor, Aug.\ Score-CAM, trails by
$0.048$ (R50) and $0.190$ (VGG). Modern gradient-free
/ optimisation CAMs (Recipro-CAM, Opti-CAM) do not fix equivariance because each still
conditions on a single orientation. The ViT gain is not an artefact of the single-view
baseline: against dedicated transformer attribution, \method{} ($0.867$) still leads
Attention Rollout~\cite{abnar2020quantifying} ($0.710$) and Transformer
Attribution~\cite{chefer} ($0.494$).

\noindent\textbf{Comparing at reduced $n$ honestly.}
\label{sec:matched}
RISE and Recipro-CAM run at $n{=}2{,}000$, and Aug.\ Score-CAM, whose inner Score-CAM costs
a forward pass per channel per view, is intractable at $10{,}000$. This is more than a
sample-size footnote: our split is assembled class-by-class ($10$ per class over $1{,}000$
classes), so a \emph{prefix} of length $k$ covers only $k/10$ classes and is an easier
sample---on the methods that did reach $10{,}000$, the first $591$ images inflate Eq by
$+0.07$ (R50 Grad-CAM $0.777$ vs.\ $0.703$) to $+0.09$ (VGG $0.599$ vs.\ $0.511$). We
therefore re-ran Aug.\ Score-CAM \emph{class-spread} (one image per class), lowering it from
$0.927$ to $0.907$ (R50) and $0.774$ to $0.766$ (VGG), and evaluated \method{} on the
\emph{identical indices} as each reduced-$n$ baseline. Both corrections \emph{widen} the CNN
gaps: against Aug.\ Score-CAM the matched margin is $0.048$ (R50, same $979$ images) and
$0.190$ (VGG), where reading across samples suggested $0.028$ and $0.183$; Recipro-CAM
trails by $0.202$ / $0.339$ and RISE by $0.544$ / $0.585$. Only ViT Recipro-CAM narrows
($0.655$ matched vs.\ $0.678$), and remains decisive. A class-spread $2{,}000$ recovers the
full population almost exactly (VGG Grad-CAM $0.513$ vs.\ $0.511$; ViT $0.259$ vs.\ $0.250$;
\method{} $0.9550$ vs.\ $0.9556$) where a prefix of the same size does not ($0.769$ vs.\
$0.703$). The bias is in the sampling, not the metric.

\begin{table}[t]
\centering\footnotesize
\caption{\textbf{Alignment, not locus, is the driver} (Eq$\uparrow$, $n{=}10{,}000$).
Identical view-averaging: \emph{without} canonical-frame alignment (row 2) it barely moves
from single-view Grad-CAM, and falls \emph{below} it on both CNNs; \emph{with} alignment it
is high in \emph{either} locus (rows 3--4), which are near-identical. The locus of
aggregation barely matters; the alignment step is decisive.}
\label{tab:align}
\setlength{\tabcolsep}{4pt}
\begin{tabular}{@{}lccc@{}}
\toprule
Aggregation variant & ResNet-50 & VGG-16 & ViT-B/16 \\
\midrule
Grad-CAM (single view)           & 0.703 & 0.511 & 0.250 \\
Unaligned output-space           & 0.690 & 0.495 & 0.317 \\
Aligned output-space             & \textbf{0.976} & \textbf{0.962} & \textbf{0.894} \\
Aligned feature-sp.\ (\method{}) & 0.956 & 0.958 & 0.867 \\
\bottomrule
\end{tabular}
\end{table}

\noindent\textbf{Alignment is the driver, not the locus.} Table~\ref{tab:align} isolates
the mechanism. Take the identical operation, average $18$ views, and toggle only
alignment. Averaging \emph{unaligned} finished maps (the route of prior multi-view
CAMs)\footnote{We aggregate unaligned views in \emph{output} space; unaligned
\emph{feature}-space averaging is geometrically ill-posed: feature tensors at different
orientations share no common frame without the very inverse-rotation that \emph{defines}
alignment, so output space is the meaningful unaligned baseline.} reaches just
$0.690/0.495/0.317$: on VGG-16 it sits \emph{below} single-view Grad-CAM ($0.511$). Adding the inverse-rotation that aligns each view to canonical lifts the same
average to $0.976$ (R50), $0.962$ (VGG) and $0.894$ (ViT). And it makes little difference \emph{where} the aligned
average is taken: aligned output-space ($0.976/0.962/0.894$) and aligned feature-space
($0.956/0.958/0.867$) differ by at most $0.027$, statistically indistinguishable for our purposes,
and both far above any single-view or unaligned baseline. This is why we frame the method
around alignment rather than the aggregation locus. We report the feature-space instantiation
as the primary one (the rest of our evaluation for faithfulness, localisation, PEUM, and the
application study is computed there), and note only that aligned output-space is marginally
higher on Eq; a practitioner may use either.

\noindent\textbf{Angle-resolved.} The gap widens with rotation and peaks near $135^\circ$,
the configuration farthest from canonical ($180^\circ$ recovers near-symmetry for many
upright objects). The supplement plots ResNet-50 and ViT-B/16 on a class-spread
$n{=}2{,}000$ sample: Grad-CAM falls to $0.558$ (R50) and $0.127$ (ViT) at $135^\circ$, while
\method{} holds $0.933$ and $0.809$ there.
\noindent\textbf{Training-free beats rotation-augmented training.} On CUB/ResNet-50,
rotation augmentation (uniform in $[-180^\circ,180^\circ]$) raises Grad-CAM equivariance
$0.82\!\to\!0.88$ while \emph{lowering} clean accuracy $0.81\!\to\!0.78$, whereas \method{}
post-hoc reaches $0.978$ on either model. We repeat this in the application domains
themselves in \S\ref{sec:application}, where the same pattern holds.

\noindent\textbf{Generality and cost.} The gain grows with backbone fragility and scale: on
ViT-L/16, Grad-CAM equivariance $0.39$ rises to $0.82$ ($+110\%$). On faithfulness,
\method{} is comparable to CAM peers: on ViT it improves \emph{both} insertion and deletion
over Grad-CAM ($0.395\!\to\!0.469$, $0.387\!\to\!0.338$), and on CNNs it is mid-pack (a small
insertion cost, e.g.\ R50 $0.587\!\to\!0.567$); we report faithfulness descriptively and do
not claim a faithfulness gain. On CUB-200-2011 localisation (a task-grounded check against ground-truth boxes, and the
kind of annotation-based metric perturbation cannot supply), \methodpcf{} gives the best
Pointing-Game score on all three backbones ($96.4/99.5/77.5\%$). Box IoU tells a more
measured story and we report it as such: ours is level with Grad-CAM on ResNet-50 ($0.607$
vs.\ $0.609$) and ahead on VGG-16 ($0.439$ vs.\ $0.411$) and ViT-B/16 ($0.274$ vs.\
$0.249$), but Grad-CAM++ and Smooth Grad-CAM++ localise better on VGG-16 ($0.536$). We
claim consistency, not localisation.

\noindent\textbf{Is aggregation too expensive?} A post-hoc operator cannot be made
equivariant without evaluating more than one view: a single forward--backward pass conditions
on exactly one orientation, and the one-pass alternative is an equivariant
\emph{architecture}~\cite{gcnn}, which requires retraining and is unavailable when auditing.
So the cost scales with $T$: measured per image on an A100, $T{=}18$ costs
$18.7$--$20.2\times$ a single Grad-CAM pass and $T{=}6$ costs $6.4$--$6.9\times$ while
retaining $93.8\%$ (ResNet-50, Eq $0.939$ of $0.955$) and $86.4\%$ (ViT-B/16, $0.784$ of
$0.868$) of the gain a full $T{=}18$ buys. \emph{We recommend $T{=}6$ as the deployable
setting},
with $T{=}18$ as the reference configuration. Alignment itself is free (unaligned averaging
at the same $T$ costs the same), and $T{=}6$ is $6$--$9\times$ cheaper than Score-CAM, which
practitioners already run. Budget is not what buys equivariance: at \method{}'s exact
wall-clock, RISE-400 reaches Eq $0.334$ and Score-CAM-80 $0.688$. The curve is steep then
flat: on ResNet-50, $T{=}2$ already recovers $44\%$ of the gain and $T{=}6$ recovers
$93.8\%$, after which $T{=}9$, $12$ and $18$ add $3.8$, $0.7$ and $1.7$ points. It is also
mildly non-monotonic, and reproducibly so on both backbones: $T{=}4$ ($0.866$ / $0.610$)
sits \emph{below} $T{=}3$ ($0.903$ / $0.646$), because view sets whose size divides $4$
place every view on the axis-aligned grid and so sample interpolation phase least
diversely. Practitioners should prefer $T\in\{3,6,9\}$ over $T{=}4$. The full cost table is
in the supplement.

%% file: sec/7_conclusion.tex
\section{Limitations and Conclusion}
\label{sec:conclusion}

\noindent\textbf{Limitations.} We evaluate in-plane \emph{rotation} only; alignment is
geometry-generic, so we expect the principle to extend, but we claim equivariance only for
rotation. Eq is an approximate empirical metric, not a guarantee. \method{} does not
universally win on faithfulness: it improves insertion \emph{and} deletion on ViT and
zero-shot CLIP but costs a little insertion on CNNs, so we report Ins/Del descriptively. The
signal-vs-noise conclusion is \emph{mostly}, not always: on prediction-flipping rotations the
model does respond differently, hence \methodpcf{} and PEUM triage. Main tables are
single-seed (per-image sign-consistency makes the ranking robust but does not bound
cross-seed variance of the mean); the domain models we trained are three-seed. PEUM is
validated as an image-level signal, not a per-pixel interval, and it ranks
\emph{explanation} instability, not model error: at predicting misclassification it is
near chance ($0.578$ AUROC) and is beaten by the prediction-flip rate ($0.764$), so it
should not be read as a correctness check. Three baselines are
compute-bound and compared on matched indices (\S\ref{sec:matched}), so their means stay less
precisely estimated. We evaluate \method{} only as a post-hoc audit; whether aligned
multi-view saliency also works as a \emph{training} signal is open.

\noindent\textbf{Responsible use.} The label-region check (\S\ref{sec:application}) shows
that \emph{no} CAM we tested, ours included, concentrates strongly on the tissue defining the
PatchCamelyon label. Rotation-consistency makes an explanation reproducible, not correct, and
a consistent-but-wrong map may be \emph{more} persuasive than a visibly unstable one. We
position \method{} and PEUM as tools that make an audit repeatable and route
\emph{irreproducible explanations} to a person, not as evidence that a model is right, and
not as a detector of when it is wrong.

\noindent\textbf{Conclusion.} Grad-CAM's rotation drift is mostly operator noise: it persists
where the representation is stable and the prediction unchanged, and that survives
conditioning on the model's own stability. \method{} removes it training-free, and a
scale-matched ablation isolates \emph{alignment}, not the locus of aggregation, as the
mechanism. It raises equivariance by $+36$ to $+247\%$ across three ImageNet backbones and
$+145\%$ on zero-shot CLIP, beats rotation-augmented training without retraining, and makes
explanations rotation-consistent in two rotation-natural domains. Code is available at
\url{https://github.com/Khawaja-Murad/EquiGrad-CAM}.